\documentclass[nohyperref]{article}

\usepackage{microtype}
\usepackage{graphicx}
\usepackage{subfigure}
\usepackage{booktabs} % for professional tables

\usepackage{hyperref}

\usepackage[accepted]{icml2022}

\usepackage{amsmath}
\usepackage{amssymb}
\usepackage{mathtools}
\usepackage{amsthm}
\usepackage{multirow}

\usepackage[capitalize,noabbrev]{cleveref}

\theoremstyle{plain}

\theoremstyle{definition}

\theoremstyle{remark}

\usepackage[textsize=tiny]{todonotes}

\begin{document}

\twocolumn[
\icmltitle{Compressed Predictive Information Coding}

% It is OKAY to include author information, even for blind
% submissions: the style file will automatically remove it for you
% unless you've provided the [accepted] option to the icml2022
% package.

% List of affiliations: The first argument should be a (short)
% identifier you will use later to specify author affiliations
% Academic affiliations should list Department, University, City, Region, Country
% Industry affiliations should list Company, City, Region, Country

% You can specify symbols, otherwise they are numbered in order.
% Ideally, you should not use this facility. Affiliations will be numbered
% in order of appearance and this is the preferred way.

% \icmlsetsymbol{equal}{*}

\begin{icmlauthorlist}
\icmlauthor{Rui Meng}{yyy}
\icmlauthor{Tianyi Luo}{ucsc}
\icmlauthor{Kristofer Bouchard}{yyy}
\end{icmlauthorlist}

\icmlaffiliation{yyy}{Lawrence Berkeley National Laboratory}
\icmlaffiliation{ucsc}{University of California, Santa Cruz}

\icmlcorrespondingauthor{Rui Meng}{rmeng@lbl.gov}

% You may provide any keywords that you
% find helpful for describing your paper; these are used to populate
% the "keywords" metadata in the PDF but will not be shown in the document
\icmlkeywords{Machine Learning, ICML}

\vskip 0.3in
]

% this must go after the closing bracket ] following \twocolumn[ ...

% This command actually creates the footnote in the first column
% listing the affiliations and the copyright notice.
% The command takes one argument, which is text to display at the start of the footnote.
% The \icmlEqualContribution command is standard text for equal contribution.
% Remove it (just {}) if you do not need this facility.

%\printAffiliationsAndNotice{}  % leave blank if no need to mention equal contribution
\printAffiliationsAndNotice{\icmlEqualContribution} % otherwise use the standard text.

\begin{abstract}
Unsupervised learning plays an important role in many fields, such as artificial intelligence, machine learning, and neuroscience. Compared to static data, methods for extracting low-dimensional structure for dynamic data are lagging. We developed a novel information-theoretic framework, Compressed Predictive Information Coding (CPIC), to extract useful representations from dynamic data. CPIC selectively projects the past (input) into a linear subspace that is predictive about the compressed data projected from the future (output). The key insight of our framework is to learn representations by minimizing the compression complexity and maximizing the predictive information in latent space. We derive variational bounds of the CPIC loss which induces the latent space to capture information that is maximally predictive. Our variational bounds are tractable by leveraging bounds of mutual information. We find that introducing stochasticity in the encoder robustly contributes to better representation. Furthermore, variational approaches perform better in mutual information estimation compared with estimates under a Gaussian assumption. We demonstrate that CPIC  is able to recover the latent space of noisy dynamical systems with low signal-to-noise ratios, and extracts features predictive of exogenous variables in neuroscience data.

\end{abstract}

\section{Introduction}
Unsupervised methods play an important role in learning robust representations that provide insight into data and exploit unlabeled data to improve performance in downstream tasks in diverse application areas \cite{bengio2013representation, chen2020simple, grill2020bootstrap, devlin2018bert,  brown2020language, baevski2020wav2vec, wang2020unsupervised}. Prior work on unsupervised representation learning can be broadly categorized into generative models such as variational autoencoders(VAEs) \cite{kingma2013auto} and generative adversarial networks (GAN) \cite{goodfellow2014generative}, and contrastive models such as contrastive predictive coding (CPC) \cite{oord2018representation} and deep autoencoding predictive components (DAPC) \cite{bai2020representation}. Generative models focus on capturing the joint distribution between representations and inputs, but are usually computationally expensive. On the other hand, contrastive models emphasize capturing the similarity structure of data in the low-dimensional representation space, and are therefore easier to scale to large datasets. 

In the case of sequential (i.e., dynamic) data, some representation learning models take advantage of an estimate of mutual information between encoded input past and output future \cite{creutzig2008predictive,creutzig2009past,oord2018representation}. Although those models provide useful representations, they are nonetheless sensitive to noise in the observational space. Dynamical Components Analysis (DCA) \cite{clark2019unsupervised} directly makes use of the mutual information between the past and the future (i.e., the predictive information \cite{bialek2001predictability}) in the latent representational space. \cite{bai2020representation} extend DCA to make use of nonlinear representations by leveraging neural networks. 

We focus on the problem of learning compressed representations of sequential data for downstream prediction tasks. A popular approach in the unsupervised representation learning literature is to use deep encoder networks to model nonlinear relations between representations and inputs \cite{chen2020simple,bai2020representation,he2020momentum}. However, use of nonlinear encoders makes interpretation of the learned features difficult. Therefore, we choose to employ a linear encoder (i.e., identify optimal linear subspaces) as in \cite{clark2019unsupervised,wang2019past}. We evaluate the quality of the representation by measuring how well the model can capture the true generative dynamics in synthetic experiment, and how well the model performs on forecasting the exogenous variables in future in real neuroscience datasets. A key innovation of our approach is to leverage a stochastic linear encoder instead of a deterministic representation. We show that stochasticity greatly improves both robustness and generalization across these tasks. 

We formalize our problem in terms of data generated from a stationary dynamical system and propose an information-theoretic objective information for CPIC that captures both compression complexity between inputs and representations as well as the predictive information in the representation space. Moreover, instead of directly estimating the objective information, we propose variational bounds and develop a tractable end-to-end training framework. We demonstrate CPIC can recover the latent trajectories of noisy dynamical systems with low signal-to-noise ratios. We conduct experiments on two neuroscience datasets, monkey motor cortex (M1) and rat dorsal hippocampus study (HC), and we show that our extracted latent representations have better forecasting capability for the monkey's future hand position for M1, and for the rat's future position for HC. The primary contributions of our paper can be summarized as follows:

\begin{itemize}
    \item We develop a novel information-theoretic framework, Compressed Predictive Information Coding (CPIC), which extracts useful representation in sequential data. It maximizes the predictive information in the representation space while minimizing the compression complexity.
    
    \item Instead of using deterministic representations, we employ a stochastic linear encoder, which contributes to better model robustness and model generalization. 
    
    \item We propose variational bounds of CPIC's objective function by taking advantage of variational bounds on mutual information and develop a tractable, end-to-end training procedure.
    
    \item We demonstrate that compared with the other unsupervised methods, CPIC recovers more accurate latent dynamics in dynamical system with low signal-to-noise ratio, and extracts more predictive features for downstream tasks in neuroscience data.
\end{itemize}

\section{Related Work}
Mutual information (MI) plays an important role in estimating the relationship between pairs of variables. It is a reparameterization-invariant measure of dependency:
\begin{align}
    I(X, Y) = \mathbb{E}_{p(x,y)} \left[\log\frac{p(x|y)}{p(x)} \right] 
\end{align}
It is widely used in computational neuroscience \cite{dimitrov2011information}, visual representation learning \cite{chen2020simple}, natural language processing \cite{oord2018representation} and bioinformatics \cite{lachmann2016aracne}. In representation learning, the mutual information between inputs and representations is used to quantify the quality of the representation and is also closely related to reconstruction error in the generative models \cite{kingma2013auto,makhzani2015adversarial}. 

Information bottleneck (IB) is a method for compressing information while retaining predictive capacity \cite{tishby2000information}. Specifically, IB compresses the variable $X$ into its compressed representation $Y$ while preserving as much information as possible related to another variable $R$. The trade-off is controlled by the trade-off parameter $\beta$ and IB is formulated as 
\begin{align}
    \min \mathcal{L}: \mathcal{L}\equiv I(X; Y) - \beta I(Y; R)\,.
\end{align}

The first term minimizes the complexity of the mapping while the second term increases the capability of the compressed data $Y$ to predict $R$. However, estimating the mutual information in the IB is computationally and statistically challenging except in two cases: discrete data, as in \cite{tishby2000information} and Gaussian data, as in \cite{chechik2005information}. However, these assumptions both severely constrain the class of learnable models \cite{alemi2016deep}. Recent works leverage deep learning models to obtain both differentiable and scalable MI estimation \cite{belghazi2018mine,nguyen2010estimating,oord2018representation,alemi2016deep,poole2019variational,cheng2020club}.

In terms of sequential data, \cite{creutzig2008predictive} utilize the IB method and develops an information-theoretic objective function. \cite{creutzig2009past} propose another IB objective function based on a specific state-space model. Recently, \cite{clark2019unsupervised} propose Dynamical Components Analysis (DCA), an unsupervised learning approach to extract a subspace with maximal dynamical structure. DCA assumes that data are stationary multivariate Gaussian process and consider a linear mapping to compress data into a low-dimensional subspace and then maximize the predictive information (PI). Due to the gaussian assumption, the mutual information can be expressed in a differentiable closed form with respect to model parameters \cite{chechik2005information}.

All of the above unsupervised representation learning models assume the data to be Gaussian, which may be not realistic, especially when applied to neuroscientific datasets \cite{o2017nonhuman,glaser2020machine}. Therefore, we leverage recently introduced lower bounds on mutual information to construct lower bounds of the CPIC objective and develop end-to-end training procedures.

\section{Compressed Predictive Information Coding}

The main intuition behind Compressed Predictive Information Coding (CPIC) is to extract a robust subspace with minimal compression complexity and maximal dynamical structure. Specifically, CPIC first discards low-level information and noise that is more local by minimizing compression complexity between inputs and representations to help model generalizations. Secondly, CPIC maximizes the predictive information in the representation space. There are different approaches to quantify the predictive capability in the literature: \cite{wiskott2002slow,turner2007maximum} target on extracting the slowly varying features; \cite{creutzig2008predictive,creutzig2009past} maximize the predictive information between past input and future input at single time stamp. Recently, \cite{clark2019unsupervised,bai2020representation} extend the input to a window size of inputs in the estimation of predictive information. Note that the predictive information based methods above are estimated based on the Gaussian assumption. CPIC relieves the Gaussian assumption by construction bounds of mutual information based on deep learning models. 

Compared with dynamical component analysis (DCA) \cite{clark2019unsupervised}, instead of employing a deterministic linear encoder to compress data, we take advantage of a stochastic linear mapping function. Given inputs, the stochastic representation follows Gaussian distributions, with the means encoded linearly while the variances encoded nonlinearly via any neural network structure. Avoiding the Gaussian assumption on mutual information \cite{creutzig2008predictive,creutzig2009past,clark2019unsupervised,bai2020representation}, CPIC leverages deep learning based estimations of mutual information. Specifically, we propose differentiable and scalable bounds of CPIC objective via variational inference, which leads to an end-to-end training. 

\begin{figure}
    \centering
    \includegraphics{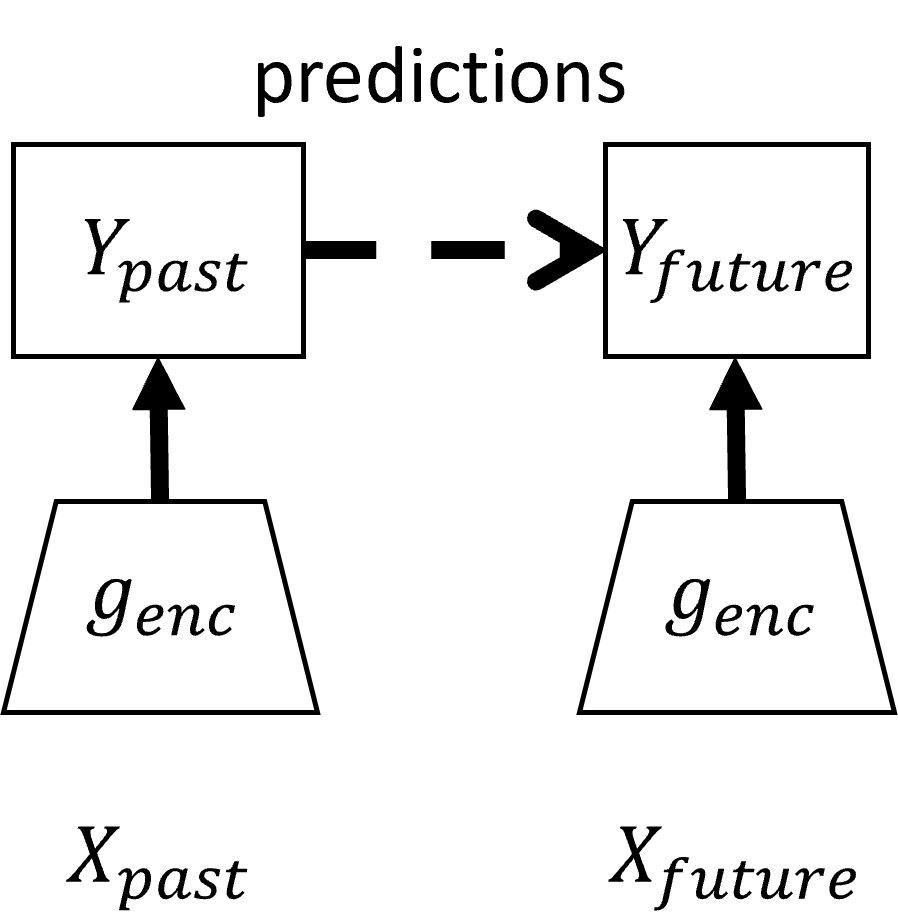}
    \caption{The overall framework of compressed predictive information coding. The encoder compress information of the input $X$ into $Y$ such that the predictive information between $Y_{\text{past}}$ and $Y_{\text{future}}$ while minimizing the mutual information between $X$ and $Y$.}
    \label{fig:CPIC}
\end{figure}

Let $X=\{x_t\}, x_t\in \mathbb{R}^N$ be a stationary, discrete time series, and let $X_{\text{past}} = (x_{-T+1}, \ldots, x_0)$ and $X_{\text{future}} = (x_1, \ldots, x_T)$ denote consecutive past and future windows of length T. Similar to the information bottleneck (IB) \cite{tishby2000information}, the CPIC objective contains a trade-off between two factors. The first seeks to minimize the compression complexity and the second to maximize the predictive information in the representation space. Both deterministic and stochastic linear encoders $p(Y|X)$ are considered. Note that when the encoder is deterministic the compression complexity is depreciated and when the encoder is stochastic the complexity is measured by the mutual information between representations and inputs. As for the predictive capability, it is measured by predictive information in representation space. In the CPIC objective, the trade-off weight $\beta>0$ dictates the balance between the compression and predictive information terms:
\begin{align}
\nonumber
\min \mathcal{L}: \mathcal{L} &\equiv \beta I(X_{\text{past}}; Y_{\text{past}}) - I(Y_{\text{past}}; Y_{\text{future}}) \label{eq:PFPC}
\end{align}
$Y_{\text{past}}$ and $Y_{\text{future}}$ are compressed versions of past and future data. Larger $\beta$ promotes a more compact mapping and thus benefits model generalization, while smaller $\beta$ would lead to more predictive information in the representation space on training data. This objective function is visualized in Figure~\ref{fig:CPIC}, where inputs $X$ are encoded into representation space as $Y$ via tractable encoders and 
the dynamics of Y are learned in a model-free manner.

The encoder $p(Y|X)$ could be implemented by fitting deep neural networks \cite{alemi2016deep} to encode data $X$. Instead, CPIC takes an approach similar to VAEs \cite{kingma2013auto}, in that it encodes data into stochastic representations. Specifically, CPIC uses a simple stochastic linear encoder ($g_{\text{enc}}$ in Figure~\ref{fig:CPIC}) to compress input $x_t$ into $y_t$ as 
\begin{align}
    y_t|x_t \sim \mathcal{N}(\mu_t, \text{diag}(\sigma_t^2))\,,
\end{align}
for each time stamp $t$. The mean of $y_t$ is given by a linear mapping $\mu_t = U^T x_t$, $U\in \mathcal{R}^{N \times D}$ whereas the variance arises from a nonlinear mapping $\sigma_t = g^{\text{Encoder}}(x_t)$. The nonlinear mapping can be modeled by any neural network architecture. For simplicity, we use a two-layer perceptron. Note that when $\sigma_t = 0$, this encoder reduces to a deterministic linear encoder. 

Given a specified window size $T$, the relation between past/future blocks of input data
$X(-T), X(T) \in \mathcal{R}^{N \times T}$ and encoded data $Y(-T), Y(T) \in \mathcal{R}^{Q \times T}$ are equivalent, $p(X(-T), Y(-T)) = p(X(T), Y(T))$, due to the stationary assumption. Without loss of generality, the compression relation can be expressed as 
\begin{align}
    Y(T) = U^T X(T) + \xi(T)
\end{align} 
where $\xi(T) \in \mathcal{N}(0, \text{blockdiag}(\text{diag}(\sigma_1^2), \ldots, \text{diag}(\sigma_T^2))$ and variance of noise $\sigma_t^2$ depends on the input $x_t$.

\subsection{Relation to Past-future Information Bottleneck}
CPIC is related to the past-future information bottleneck (PFIB) in \cite{creutzig2008predictive, creutzig2009past}. There are two primary differences. First, PFIB only compresses the past information, while CPIC compresses both the past and the future. The intuition behind this is similar to that in \cite{oord2018representation}: we want to discard low-level information and localized noise by maximizing the mutual information between the two in the representation space. By using a linear encoder, CPIC eases interpretation of optimal subspace with minimal compression complexity and maximal dynamical structure. Second, PFIB makes a Gaussian assumption \cite{chechik2005information} in terms of estimation of mutual information, while CPIC does not make any assumption on the data. Recently, \cite{wang2019past} utilize the PFIB framework to learn the minimally complex yet most predictive representation. That work has two primary drawbacks. It considers a deterministic linear encoder, directly ignoring the effect of complexity in compression, and it directly utilizes a neural network to build the relation between past and future, which may have a profound effect on the expressiveness of the model. 

\section{Variational Bounds of Compressed Predictive Information Coding}

In the CPIC, since data $X$ are stationary, the mutual information for the past is equivalent to that for the future $I(X(-T); Y(-T)) = I(X(T); Y(T))$. Therefore, the objective of CPIC can be rewritten as 
\begin{align}
    \min \mathcal{L} = \beta I(X(T); Y(T)) - I(Y(-T); Y(T))\,. \label{eq:PFPC_T}
\end{align}
We develope the variational upper bounds on the mutual information for the compression complexity $I(X(T); Y(T))$ and lower bounds on mutual information for the predictive capacity $I(Y(-T); Y(T))$.

\subsection{Upper Bounds of Compression Complexity}
Directly estimating compression complexity $I(X(T); Y(T)) := \mathbb{E}_{X(T)}\big[\text{KL}(p(y(T)| x(T)), p(y(T)))\big]$ is intractable, because the population distribution $p(y(T))$ is unknown. Thus we introduce a variational approximation to the marginal distribution of encoded inputs $p(y(T))$, denoted as $r(y(T))$. Due to the non-negativity of the Kullback-Leibler (KL) divergence, we can derive the variational upper bound as 
\begin{align}
    I(X(T); Y(T)) & = \mathbb{E}_{X(T)}\big[ \text{KL}(p(y(T)|x(T)), r(y(T))) \nonumber\\ 
                  &- \text{KL}(p(y(T)), r(y(T))) \big]\nonumber \\ 
                  & \leq \mathbb{E}_{X(T)}\big[ \text{KL}(p(y(T)|x(T)), r(y(T))) \big] \nonumber \\
                  & = I_{\text{VUB}}(X(T); Y(T)) \,. \label{eq:VUB}
\end{align}

Generally, learning $r(y(T))$ was recognised as the distribution density estimation problem \cite{silverman2018density}, which is challenging. In this setting, the variational distribution $r(y(T))$ is assumed to be learnable, and thus estimating the variational upper bound is tractable. In particular, \citet{alemi2016deep} fix $r(y(T))$ as a standard normal distribution, leading to high-bias in MI estimation. Recently, \citet{poole2019variational} replace $r(y(T))$ with a Monte Carlo approximation. Specifically, with $S$ sample pairs $(x(T)_i, y(T)_i)_{i=1}^S$, $r_i(y(T)) = \frac{1}{S-1}\sum_{j\neq i} p(y(T)|x(T)_j) \approx p(y(T))$ and \citet{poole2019variational} derive a leave-one-out upper bound (L1Out) as below:
\begin{align}
    & I_{\text{L1Out}}(X(T); Y(T)) \nonumber \\
    & = \mathbb{E}\left[\frac{1}{S} \sum_{i=1}^S\left[ \log \frac{p(y(T)_i| x(T)_i)}{\frac{1}{S-1}\sum_{j\neq i}p(y(T)_i | x(T)_j)}\right] \right]\,.  \label{eq:L1Out}
\end{align}

In practice, L1Out bound depends on the sample size $S$ and may suffer from numerical instability. Thus, we would like to choose the sample size $S$ as large as possible. In general scenarios where $p(y(T)|x(T))$ is intractable, \citet{cheng2020club} propose the variational versions of VUB and L1Out by using a neural network to approximate the condition distribution $p(y(T)|x(T))$ in \eqref{eq:VUB} and \eqref{eq:L1Out}. Since CPIC leverages a known stochastic/deterministic linear encoder, VUB and L1Out estimators are considered in the construction of CPIC variational bounds.

\subsection{Lower Bounds of Predictive Information}
As for the predictive information (PI), we would derive lower bounds of $I(Y(-T); Y(T))$ using results in \cite{agakov2004algorithm,alemi2016deep,poole2019variational}. First, similar to \citet{agakov2004algorithm}, we replace the intractable conditional distribution $p(y(T)|y(-T))$ with a tractable optimization problem over a variational conditional distribution $q(y(T)|y(-T))$. It yields a lower bound on PI due to the non-negativity of the KL divergence:
\begin{align}
    I(Y(-T); Y(T)) & \geq H(Y(T)) \nonumber \\
    & + \mathbb{E}_{p(y(-T), y(T))}[\log q(y(T)|y(-T))] \label{eq:lowerbound}
\end{align}
where $H(Y)$ is the differential entropy of variable $Y$ and this bound is tight if and only if $q(y(T)|y(-T)) = p(y(T)|y(-T))$, suggesting that the second term in \eqref{eq:lowerbound} equals the negative conditional entropy $ - H(Y(T)|Y(-T))$. 

Because $H(Y(T)) \geq 0$ has no information about the dependence between $Y(-T)$ and $Y(T)$, it leads to a looser lower bound by ignoring it as the same in \cite{agakov2004algorithm} such that
\begin{align}
    I(Y(-T), Y(T)) & \geq \int \mathbb{E}_{p(y(-T), y(T))}[\log q(y(T)|y(-T))] \nonumber \\
    & \triangleq I_{\text{LBA}}(Y(-T), Y(T))\,. \label{eq:LBA} 
\end{align}

The conditional expectation in \eqref{eq:LBA} can be estimated using Monte Carlo sampling based on the encoded data distribution $p(y(-T), y(T))$. And encoded data are sampled by introducing the augmented data $x(-T)$ and $x(T)$ and marginalizing them out as 
\begin{align}
    p(y((-T), y(T)) & = \int p(x(-T), x(T))p(y(-T)|x(-T)) \nonumber \\
    & p(y(T)|x(T)) dx(-T)x(T) \label{eq:latent_sampling}
\end{align} 
according to the Markov chain proposed in Figure~\ref{fig:CPIC}. 

However, this lower bound requires a tractable decoder for the conditional distribution $q(y|x)$ \cite{alemi2016deep}. Alternatively, according to \cite{poole2019variational}, by considering an energy-based variational family to express and conditional distribution $q(y(T)|y(-T))$:
\begin{align}
    q(y(T)|y(-T)) = \frac{p(y(T))e^{f(y(T), y(-T))}}{Z(y(-T))} \nonumber
\end{align}
where $f(x,y)$ is a differentiable critic function, $Z(y(-T)) = \mathbb{E}_{p(y(T)}\left[ e^{f(y(T), y(-T))} \right]$ is a partition function, and introducing a baseline function $a(y(T))$, we derive a tractable unnormalized Barber and Agakov (TUBA) lower bound of the predictive information as:
\begin{align}
    I(Y(-T), Y(T)) & \geq \mathbb{E}_{p(y(-T), y(T))}[\tilde{f}(y(-T), y(T))] \nonumber \\
    & - \log\left( \mathbb{E}_{p(y(-T))p(y(T))}[e^{\tilde{f}(y(-T), y(T))}]\right) \nonumber \\
    & \triangleq I_{\text{TUBA}}(Y(-T), Y(T)) \label{eq:TUBA}
\end{align}
where $\tilde{f}(y(-T), y(T)) = f(y(-T), y(T)) - \log(a(y(T)))$ is treated as an updated critic function. Notice that different choices of baseline functions lead to different mutual information estimators. When $a(y(T)) = 1$, it leads to mutual information neural estimator (MINE) \cite{belghazi2018mine}; when $a(y(T)) = Z(y(T))$, it leads to the lower bound proposed in \cite{donsker1975asymptotic} (DV) and when $a(y(T)) = e$, it recovers the lower bound in \cite{nguyen2010estimating} (NWJ) also known as f-GAN \cite{nowozin2016f} and MINE-f \cite{belghazi2018mine}. In general, the critic function $f(x,y)$ and the log baseline function $a(y)$ are usually parameterized by neural networks \cite{oord2018representation,belghazi2018mine}. Especially, \citet{oord2018representation} use a separable critic function $f(x,y) = h_\theta(x)^T g_\theta(y)$  while \citet{belghazi2018mine} use a joint critic function $f(x,y) = f_\theta(x, y)$. \cite{poole2019variational} claim that joint critic function generally performs better than separable critic function but scale poorly with batch size. 

\begin{figure*}[ht!]
    \centering
    \includegraphics[width=\textwidth]{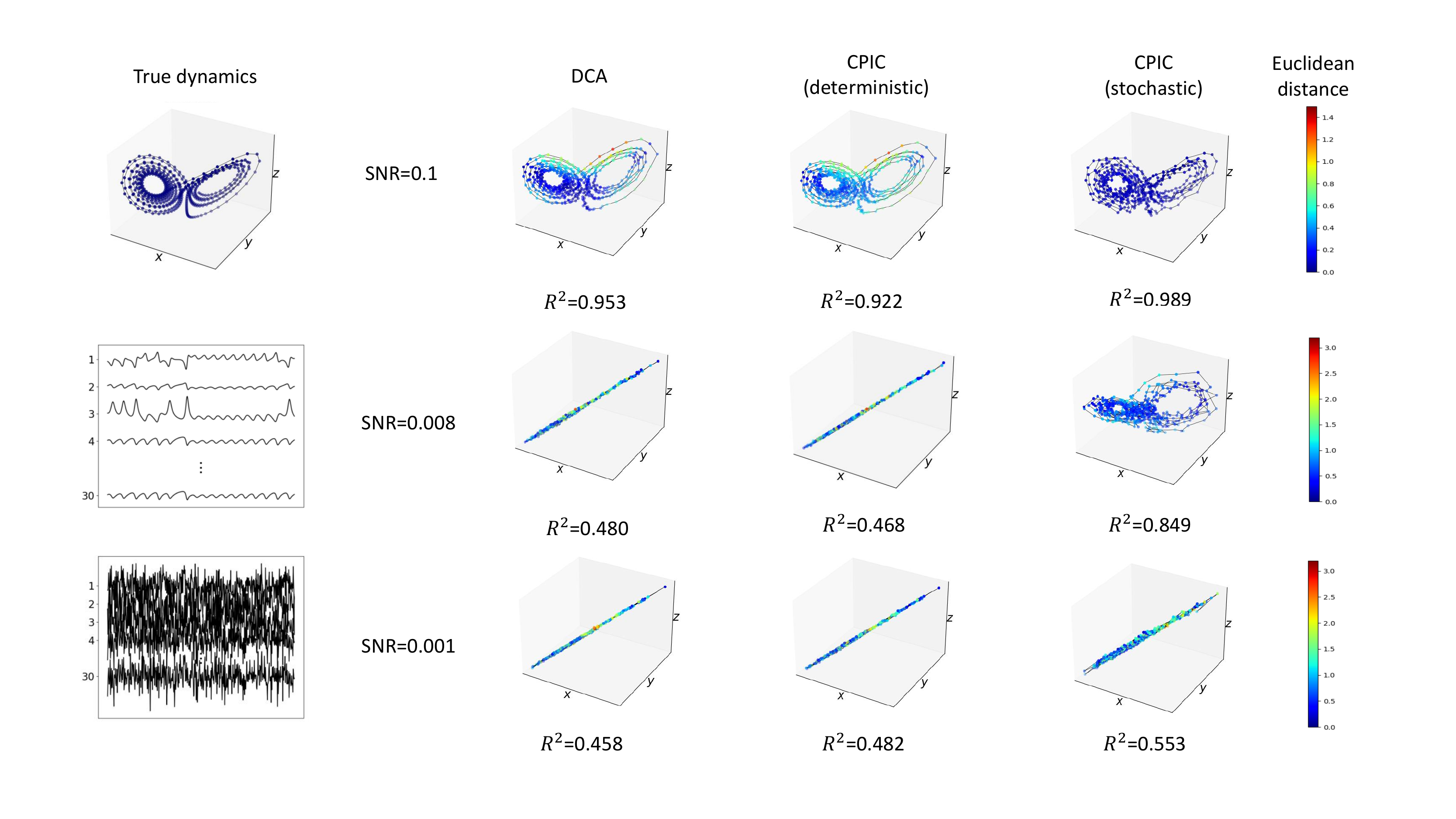}
    \caption{\textbf{Left panel.} Top: 3D trajectories of lorenz attractor's ground-truth. Middle: 30D projected trajectory. Bottom: Corrupted 30D trajectory with SNR=0.001. \textbf{Right Panel.} 3D trajectories obtained by DCA, deterministic CPIC, and stochastic CPIC methods with respect to different SNRs (0.1, 0.008, 0.001). We encode the point-wise Euclidean distance between the aligned inferred latent 
   dynamics and the true dynamics into color on trajectories. Color from blue to red corresponds to the distance from short to long respectively. Separate colorbars are used for their corresponding SNRs. Stochastic CPIC outperforms DCA and deterministic CPIC with more similar dynamics pattern and smaller point-wise errors.}
    \label{fig:latent_rep}
\end{figure*}

However, all MI estimators in the form of \eqref{eq:TUBA} have high variance due to the high variance of $f(x,y)$. \citet{oord2018representation} propose a low-variance MI estimator based on noise-contrastive estimation called InfoNCE. In our case, the lower bound of predictive information is derived as
\begin{align}
    I(Y(-T); Y(T)) \geq \mathbb{E}\left[ \frac{1}{S}\sum_{i=1}^S \log \frac{e^{f(y(-T)_i, y(T)_i)}}{\frac{1}{S} \sum_{j=1}^S e^{f(y(-T)_i, y(T)_j)}} \right] \,
\end{align}

The expectation is over $S$ independent samples from the joint distribution: $p(y(-T), y(T))$ with the same sampling procedure in \eqref{eq:latent_sampling}.

\subsection{Variational Bounds of CPIC}
We propose two classes of lower bounds of CPIC based on whether the bounds depend on multiple samples. We name the first class as Uni-sample lower bounds, which take the VUB upper bound of mutual information for the complexity of data compression $I(X(T), Y(T))$ and the TUBA as the lower bound of predictive information in \eqref{eq:TUBA}. Thus we have
\begin{align}
    \mathcal{L}_{\text{UNI}} &  = \beta \text{KL}(p(y(T)|x(T)), r(y(T))) \nonumber \\
    & -  I_{\text{TUBA}}(Y(-T), Y(T))\,. \label{eq:uni_sample_lower_bounds}
\end{align}

Notice that by choosing different baseline functions, the lower bounds are equivalent to different mutual information estimators such as MINE, DV and NWJ. The second class is named as multiple-sample lower bounds, which take advantage of the noise-contrastive estimation trick. And the multi-sample lower bounds are expressed as
\begin{align}
    \mathcal{L}_{\text{MUL}} & = \beta \mathbb{E}\left[\frac{1}{S} \sum_{i=1}^S \left[ \log \frac{p(y(T)_i| x(T)_i)}{\frac{1}{S-1}\sum_{j\neq i}p(y(T)_i | x(T)_j)}\right]\right] \nonumber \\
    & - \mathbb{E}\left[ \frac{1}{S}\sum_{i=1}^S \log \frac{e^{f(y(-T)_i, y(T)_i)}}{\frac{1}{S} \sum_{j=1}^S e^{f(y(-T)_i, y(T)_j)}} \right]\,.
    \label{eq:multiple_sample_lower_bounds}
\end{align}

Two main differences exist between these two classes of lower bounds. First, the performance of multiple-sample lower bounds depend on batch size while uni-sample lower bounds do not. Secondly, multiple-sample lower bounds have lower variance than uni-sample lower bounds. Without any specification, we assume that CPIC employs the multiple-sample lower bound.

On the other hand, we classify the lower bounds into stochastic and deterministic version by whether we employ a deterministic or stochastic encoder. Notice that when choosing the deterministic encoder, the compression complexity term (first term) in \eqref{eq:uni_sample_lower_bounds} and \eqref{eq:multiple_sample_lower_bounds} are constant. 

\section{Experiments}

In this section, we illustrate the performance of CPIC in both synthetic and real data experiments. We first show the reconstruction performance of CPIC in noisy dynamical system on one tractable experiment (noisy Lorenz attractor). We demonstrate that CPIC recovers more accurate latent dynamics compared with other unsupervised methods. Secondly, for further demonstrating the benefits of the latent representations, we conduct forecasting of exogenous variables using simple linear models from neuroscience data sets. The motivation for using linear forecasting models is that good representations contribute to disentangling complex data in a linearly accessible way. Specifically, we extract latent representations and then conduct forecasting tasks given the inferred representations on two neuroscience datasets: multi-neuronal recordings from the hippocampus (HC) while rats navigate a maze \cite{glaser2020machine} and muli-neuroinal recordings from primary motor cortex (M1) during a reaching task for monkeys \cite{o2017nonhuman}. The experimental results show that CPIC has better predictive performance on those forecasting tasks comparing with existing methods.

\subsection{Noisy Lorenz Attractor}

The Lorenz attractor is a 3D time series that are realizations of the Lorenz dynamical system \cite{pchelintsev2014numerical}. It describes a two dimensional flow of fluids with latent processes given as:
\begin{align}
    \frac{df_1}{d t} & = \sigma(f_2 - f_1), \frac{d f_2}{d t}=f_1(\rho-f_3) - f_2, \nonumber \\
    \frac{df_3}{dt} & = f_1 y- \beta f_3\,. \label{eq:lorenz}
\end{align}
Lorenz sets the values $\sigma=10, \beta=8/3$ and $\rho=28$ to exhibit a chaotic behavior, as done in recent works \cite{she2020neural,clark2019unsupervised,zhao2017variational,linderman2017bayesian}. We simulate the latent signals from the Lorenz dynamical system and show them in the left-top panel in Figure~\ref{fig:latent_rep}. Then we map the 3D latent signals to 30D lifted observations with a random linear embedding in the left-middle panel and add spatially anisotropic Gaussian noise on the 30D lifted observations in the left-bottom panel. The noises are generated according to different signal-to-noise ratios (SNRs), where SNR is defined by the ratio of the variance of the first principle components of dynamics and noise as in \cite{clark2019unsupervised}. Specifically, we propose 10 different SNR levels spaced evenly on a log (base 10) scale between [-3, -1] and corrupt the 30D lifted observations with noise corresponding to different SNR levels. Finally, we deploy different representation learning methods to recover the true 3D dynamics from different corrupted 30D lifted observations with different SNR levels, and compare the accuracy of recovering the underlying Lorenz attractor time-series. 

We first compare stochastic CPIC with deterministic CPIC and the DCA method. For both CPIC methods, we take multiple-sample lower bounds $\mathcal{L}_{\text{MUL}}$. We set the latent dimension size $Q=3$ and the time window size $T=4$. We aligned the inferred latent trajectory with the true 3D dynamics with optimal linear mapping due to the reparameterization-invariant measure of latent trajectories. We validate the reconstruction performance based on the $R^2$ regression score. For all three methods, aligned latent trajectories inferred from corrupted lifted observation with different $\text{SNR}=0.001, 0.008$ and $0.1$ levels of noise as well as $R^2$ scores are shown in Figure~\ref{fig:latent_rep}. The point-wise distances between the recovered dynamics and the ground-truth dynamics are encoded into different colors from blue to red, representing the distance from short to long. 
For high SNR (SNR = 0.1, top-right), all methods did a good job of recovering the Lorenz dynamics and the stochastic CPIC had larger $R^{2}$ than both DCA and CPIC. For intermediate SNR (SNR = 0.008, middle-right), we see that DCA and deterministic CPIC completely fail at recovering the ground truth dynamics, while stochastic CPIC performs reasonably. Finally, as the SNR gets lower (SNR = 0.001, bottom-right) all methods perform poorly, but stochastic CPIC has higher $R^2$ scores than others. These results suggest that stochastic CPIC outperforms its deterministic version and DCA across different SNRs. 

\begin{figure}[ht!]
    \centering
    \includegraphics[width=0.5\textwidth,height=0.35\textwidth]{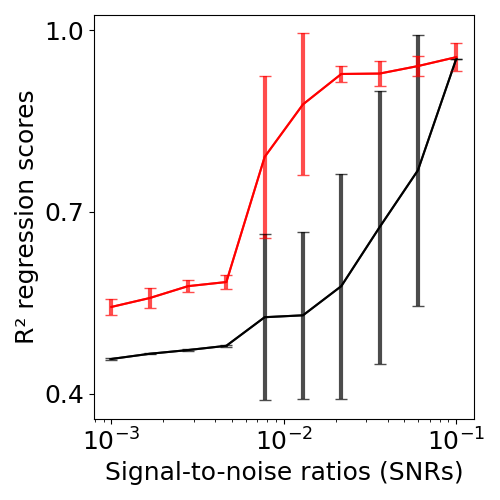}
    \caption{Comparison on $R^2$ scores of latent dynamics regression between DCA and stochastic CPIC with the mean and one standard deviation above/below it over running ten different random intializations.}
    \label{fig:real_data_uq}
\end{figure}

To better illustrate the latent dynamics recovery in terms of different SNRs, we plot the $R^2$ scores from stochastic CPIC and DCA method for all ten SNR levels with uncertainty quantification in Figure~\ref{fig:real_data_uq}. The uncertainty quantification is visualized by one standard deviation below and above and mean of ten $R^2$ scores with respect to their corresponding ten random initializations. As suggested by the examples in Figure~\ref{fig:latent_rep}, it shows that stochastic CPIC robustly outperforms DCA in recovering latent dynamics, and in particular, the SNR at which reasonable recovery is lower for stochastic CPIC than DCA.  In addition, we conducted another experiment to compare the performance of uni-sample lower bounds and multi-sample lower bounds between the deterministic and stochastic versions of CPIC. The CPIC with uni-sample lower bounds refers to those with NWJ, MINE, and TUBA estimates of predictive information (P). The CPIC with multiple-sample lower bound refers to that with NEC estimate of PI. We report the performance among these four variants of objectives in both deterministic and stochastic CPICs in appendix (Table~\ref{tab:R2_lorenz}). We find that stochastic CPIC with multiple-sample lower bounds has better reconstruction performance of latent dynamics in majority levels of noisy data. 

\begin{figure*}[ht!]
    \centering
    \includegraphics[width=\textwidth]{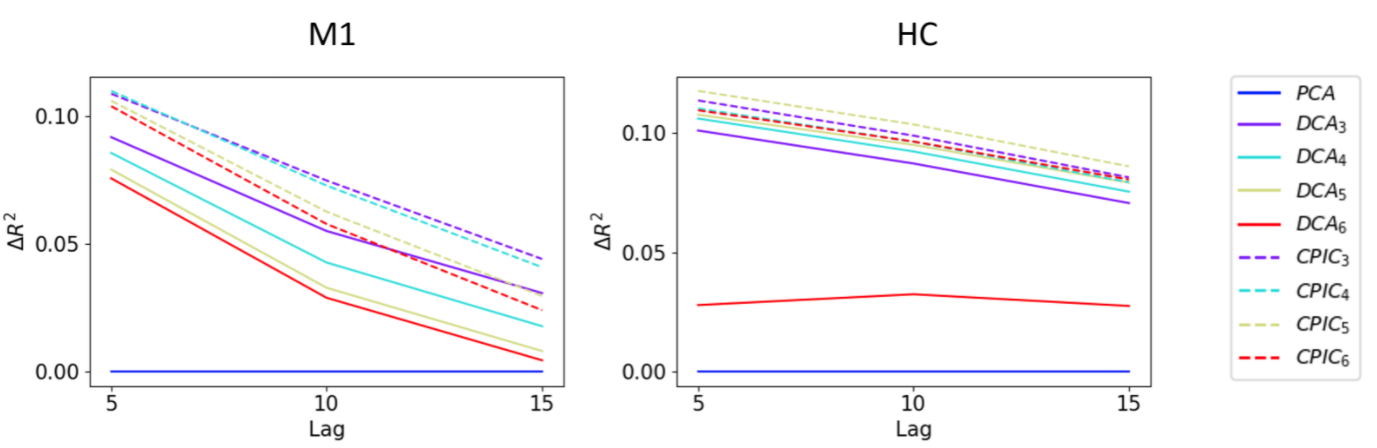}
    \caption{Comparison between different methods (PCA, DCA, and stochastic CPIC) on $R^2$ regression scores for two forecasting tasks with three different lag values (5, 10, and 15). $R^2$ values are averaged across five folds. For DCA and stochastic CPIC models using predictive information loss, we test them with four time window sizes (T = 3, 4, 5, 6). The subscript refers to the window size of past and future time series. Left: motor cortex (M1). Right: dorsal hippocampus (HC).}
    \label{fig:real_data_R2}
\end{figure*}

% As for the generating details of true 3D lorenz attractor and noisy 30D trajectories, we simulated $Q=3$-dimensional state trajectories $f^{(q)}, q=1,2,3$ via the Lorenz system described by a two-dimensional flow of fluids: 
% \begin{align}
%     \frac{df^{(1)}}{d t} & = \sigma(f^{(2)} - f^{(1))}, \nonumber \\
%     \frac{d f^{(2)}}{d t}&=f^{(1)}(\rho-f^{(3)}) - f^{(2)}, \nonumber \\
%     \frac{df^{(3)}}{dt} & = f^{(1)} y- \beta f^{(3)}\,. \label{eq:lorenz}
% \end{align}
% We set hyper-parameters $\sigma=10, \beta=8/3$ and $\rho=28$ to exhibit a chaotic behavior as the same utilized in recent works \cite{she2020neural,linderman2017bayesian,meng2021bayesian}. The state trajectories are normalized in each dimension with unit variance and zero mean. Then we randomly generate data via a stochastic linear mapping: as for $t$th observation, $y_t = U^T f_t + \xi_t$ where $U \in \mathbb{R}^{P \times Q}$ is a random semi-orthogonal matrix and random noise $\xi_t$ are generated weakly orthogonal to signal subspace following the same procedures in \cite{clark2019unsupervised}. 

\begin{table*}[ht!]
    \centering
    \begin{tabular}{c|c|c|cc|cc}
        \hline
        \multirow{2}{*}{Dataset} & \multirow{2}{*}{Lag} & \multirow{2}{*}{DCA} & \multicolumn{4}{c}{CPIC} \\
        \cline{4-7}
        & & & Deterministic & Improvement over DCA & Stochastic & Improvement over DCA \\
        \hline
        \multirow{3}{*}{M1} & 5 & 0.251 & 0.265 & 5.58\% & 0.268 & 6.77\% \\
        & 10 & 0.259 & 0.275 & 6.18\% & 0.279 & 7.72\% \\
        & 15 & 0.210 & 0.222 & 5.71\% & 0.224 & 6.67\% \\
        \hline
        \multirow{3}{*}{HC} & 5 & 0.149 & 0.147 & -1.34\% & 0.159 & 6.71\% \\
        & 10 & 0.132 & 0.131 & 0.76\% & 0.141 & 6.82\% \\
        & 15 & 0.111 & 0.110 & -0.90\% & 0.118 & 6.31\% \\
        \hline
    \end{tabular}
    \caption{Comparison between DCA, deterministic CPIC and stochastic CPIC on $R^2$ regression scores on M1 and HC datasets with the optimal window size among $T \in [3,4,5,6]$ for three different lag values (5, 10, and 15). $R^2$ regression scores are averaged across five folds. We report the deterministic CPIC's and stochastic CPIC's improvement (percentage) over DCA and it shows that stochastic CPIC robustly improves the forecasting performance on downstream tasks.}
    \label{tab:real_data_R2}
\end{table*}

% Firstly, we show the recovered 3D trajectories of using DCA, deterministic PFPC, and stochastic PFPC methods for different SNR parameters (0.1, 0.01, 0.001) in Figure \ref{fig:latent_rep}. We also listed the R2 score between the true latent trials and optimal inferred latent trials obtain by the optimal linear transform. From Figure \ref{fig:latent_rep}, we can observe that our proposed PFPC methods can obtain better 3D trajectories especially in a highly signal noise ratio which demonstrates the stronger model robustness of PFPC. Comparing between deterministic and stochastic PFPC, we find that stochastic PFPC performs better. It meets our expectation because stochastic PFPC considers the variance of parameters in the latent subspace the and are more suitable for the noise situation. In Figure \ref{fig:PFPC}, we show the comparison on R2 score between DCA and stochastic PFPC for all 10 different SNR numbers.  

\subsection{Forecasting Exogenous Variables from Multi-neuronal Recordings with Linear Decoding}

In this section, we show that latent representations extracted by stochastic CPIC perform better in the downstream forecasting tasks on two neuroscience datasets: multi-neuron recordings from monkey motor cortex (M1) during a reaching task \cite{o2017nonhuman} and multi-neuron recordings from rat dorsal hippocampus (HC) during maze navigation \cite{glaser2020machine}. We compared stochastic CPIC with PCA and DCA. For each model, we extract the latent representations and conduct prediction tasks on the relevant exogenous variable at a future time step. For example, for the M1 dataset, we extract a consecutive 3-length window representation of multi-neuronal spiking activity to predict the monkey's arm position in a future time step which is $lag$ time stamps away. The predictions are conducted by linear regression \footnote{https://scikit-learn.org/stable/modules/linear\_model.html} to emphasize the structure learned by the unsupervised methods. For these tasks, $R^2$ regression score is used as the evaluation metric to measure the forecasting performance. $R^2$ regression scores were calculated using 5-fold cross validation. We considered three different lag values (5, 10, and 15), and considered four different window sizes $T=3,4,5,6$ in both DCA and stochastic CPIC. 

The results of this analysis are visualized in Figure~\ref{fig:real_data_R2}. Generally, a larger $lag$ value leads to a more challenging forecasting task. Importantly, we observe that models utilizing the predictive information loss perform much better than PCA, and moreover, stochastic CPIC leads to consistently better prediction performance than DCA on any fixed window size $T$ for all forecasting tasks. This demonstrates that stochastic CPIC extracts more useful information than DCA in these complex neuroscience datasets. Moreover, we also find that different settings of window size impact the $R^2$ regression score. M1 obtains the best $R^2$ when the window size is set to 3. For HC, it is 5. Similar to DCA, the optimal window size of past/future data depends on the characteristics of datasets as well as the specified downstream tasks. Finally, we show the necessity of the stochastic encoder by comparing deterministic CPIC with stochastic CPIC on M1 and HC datasets on all lag values. We choose the optimal window size $T\in [3,4,5,6]$ for stochastic CPIC, which is $T=3$ for M1 and $T=5$ for HC. The $R^2$ scores for DCA, deterministic CPIC and stochastic CPIC are shown in Table~\ref{tab:real_data_R2} on their improvement over DCA. Together, these results suggest that stochastic CPIC robustly improves the predictive performance on downstream tasks comparing to DCA and deterministic CPIC. 

% In the unusual situation where you want a paper to appear in the
% references without citing it in the main text, use \nocite
% \nocite{langley00}

\section{Concluding Remarks}
%Unsupervised learning plays an important role in artificial intelligence. We develop a novel information-theoretic framework, Past-future Predictive Coding (PFPC), to extract useful representation in sequential data, which selectively projects input past into a linear subspace that is predictive about the compressed data projected from the output future. The key insight of our model is to learn representations by minimizing the compression complexity and maximizing the predictive information in latent space. We propose variational bounds of PFPC loss which induce the latent space to capture information that is maximally predictive. Our variation bounds are tractable by leveraging bounds of mutual information. We find that stochasticity of encoder robustly contributes to better representation and variational approaches perform better in mutual information estimation compared with that under a Gaussian assumption. We demonstrate that our approach is able to recover the latent space of noisy dynamical systems with highly signal noise ratio and extract predictive features achieving strong performance on downstream tasks in neuron data.

We developed a novel information-theoretic framework, Compressed Predictive Information Coding, to extract useful representations in sequential data. CPIC maximizes the predictive information in representation space while minimizing the compression complexity. We leverage stochastic representations by employing a stochastic linear encoder and propose variational bounds of the CPIC objective function. We demonstrate that CPIC can extract more accurate low-dimensional latent dynamics and more useful representations that has better forecasting performance of downstream tasks on two neuroscience datasets. Together, these results suggest that CPIC will yield similar improvements in other datasets.

%\newpage
\clearpage

\bibliography{example_paper}
\bibliographystyle{icml2022}

%%%%%%%%%%%%%%%%%%%%%%%%%%%%%%%%%%%%%%%%%%%%%%%%%%%%%%%%%%%%%%%%%%%%%%%%%%%%%%%
%%%%%%%%%%%%%%%%%%%%%%%%%%%%%%%%%%%%%%%%%%%%%%%%%%%%%%%%%%%%%%%%%%%%%%%%%%%%%%%
% APPENDIX
%%%%%%%%%%%%%%%%%%%%%%%%%%%%%%%%%%%%%%%%%%%%%%%%%%%%%%%%%%%%%%%%%%%%%%%%%%%%%%%
%%%%%%%%%%%%%%%%%%%%%%%%%%%%%%%%%%%%%%%%%%%%%%%%%%%%%%%%%%%%%%%%%%%%%%%%%%%%%%%

\appendix
\onecolumn

\section{Comparison between DCA, deterministic \& stochastic CIPC (four different variational lower bounds) on $R^2$ regression score}

In this section, the $R^2$ regression scores for DCA, deterministic \& stochastic CIPC (NWJ, MINE, TUBA, and NEC lower bounds) for all ten different SNRs are reported in Table~\ref{tab:R2_lorenz}. Generally, It shows that stochastic CPIC with multiple-sample lower bound outperforms other approaches in majority of SNRs.

\begin{table*}[ht!]
    \centering
    \begin{tabular}{c|c|c|c|c|c|c|c|c|c}
        \hline
          \multirow{3}{*}{} 
          &  \multirow{3}{*}{} 
          & \multicolumn{8}{c}{CIPC} \\
          \cline{3-10}
          & &  \multicolumn{4}{c|}{deterministic} & \multicolumn{4}{c}{stochastic} \\
          %\cline{3-10}
          Signal Noise Ratio & DCA &  NWJ & MINE & TUBA & NEC & NWJ & MINE & TUBA & NEC \\
        \hline  \hline
        SNR = 0.001   & 0.458 & \textbf{0.554} & 0.543 & 0.547 & 0.482 & 0.539 & 0.550 & 0.553 & 0.459 \\
        \hline  
        SNR = 0.00167 & 0.466 & 0.539 & 0.538 & 0.574 & 0.430 & 0.573 & 0.569 & 0.571 & \textbf{0.576} \\
        \hline  
        SNR = 0.00278 & 0.473 & 0.573 & 0.573 & 0.573 & 0.413 & 0.587 & 0.583 & \textbf{0.590} & 0.588 \\
        \hline  
        SNR = 0.00464 & 0.478 & 0.579 & 0.562 & 0.584 & 0.438 & \textbf{0.598} & 0.583 & 0.556 & 0.593 \\
        \hline  
        SNR = 0.00774 & 0.480 & 0.597 & 0.559 & 0.515 & \textbf{0.912} & 0.582 & 0.579 & 0.589 & 0.598 \\
        \hline  
        SNR = 0.01292 & 0.484 & 0.587 & 0.596 & 0.597 & 0.468 & 0.580 & 0.563 & 0.592 & \textbf{0.923} \\
        \hline  
        SNR = 0.02154 & 0.486 & 0.590 & 0.596 & 0.592 & 0.688 & 0.568 & 0.599 & 0.864 & \textbf{0.930} \\
        \hline  
        SNR = 0.03594 & 0.491 & 0.587 & 0.912 & 0.594 & 0.923 & 0.937 & 0.632 & 0.907 & \textbf{0.951} \\
        \hline  
        SNR = 0.05995 & 0.952 & 0.933 & 0.837 & 0.936 & 0.474 & 0.970 & 0.939 & 0.896 & \textbf{0.970} \\
        \hline  
        SNR = 0.1     & 0.953 & 0.920 & 0.893 & 0.889 & 0.922 & 0.926 & 0.910 & 0.854 & \textbf{0.989} \\
        \hline
    \end{tabular}
    \caption{$R^2$ regression scores for DCA, deterministic \& stochastic CIPC (NWJ, MINE, TUBA, and NEC lower bounds) for all ten different SNRs %Lorenz reconstruction performance (R2) as a function of the SNR for all methods.
    }
    \label{tab:R2_lorenz}
\end{table*}

%%%%%%%%%%%%%%%%%%%%%%%%%%%%%%%%%%%%%%%%%%%%%%%%%%%%%%%%%%%%%%%%%%%%%%%%%%%%%%%
%%%%%%%%%%%%%%%%%%%%%%%%%%%%%%%%%%%%%%%%%%%%%%%%%%%%%%%%%%%%%%%%%%%%%%%%%%%%%%%

\end{document}